\documentclass[conference]{IEEEtran}
\IEEEoverridecommandlockouts

\usepackage{cite}
\usepackage{amsmath,amssymb,amsfonts}
\usepackage{algorithmic}
\usepackage{graphicx}
\usepackage{textcomp}
\usepackage{xcolor}
\usepackage{eso-pic} 

\usepackage{booktabs}
\usepackage{tabularx}
\usepackage{multirow}
\usepackage{array}
\usepackage{graphicx}

\def\BibTeX{{\rm B\kern-.05em{\sc i\kern-.025em b}\kern-.08em
    T\kern-.1667em\lower.7ex\hbox{E}\kern-.125emX}}
\begin{document}

\AddToShipoutPictureFG*{%
  \AtPageLowerLeft{%
    \raisebox{45pt}[0pt][0pt]{%
      \hspace*{55pt}%
      \parbox{500pt}{\raggedright\fontsize{7.5pt}{8.8pt}\selectfont
        \textcopyright{} 2026 IEEE. Personal use of this material is permitted.
        Permission from IEEE must be obtained for all other uses, in any current or future media,
        including reprinting/republishing this material for advertising or promotional purposes,
        creating new collective works, for resale or redistribution to servers or lists, or reuse
        of any copyrighted component of this work in other works.
      }%
    }%
  }%
}

\title{Conditional Diffusion-Based Proxy Representation Recovery for BCVA Prediction with Missing SLO Modality
 \thanks{This work was supported in part by the Key Research and Development Program of Hubei Province (No. 2025BAB020) and the China University Industry-Academia-Research Innovation Fund (No. 2025XJS023). Corresponding author: Hua Zou (zouhua@whu.edu.cn).}
}

 \author{ \IEEEauthorblockN{ Chuyi He\IEEEauthorrefmark{1}, Hua Zou\IEEEauthorrefmark{2}, Fei Luo\IEEEauthorrefmark{2}, Zixian Yang\IEEEauthorrefmark{2} } \IEEEauthorblockA{ \IEEEauthorrefmark{1}School of Electrical Engineering, Northeast Electric Power University, China\\ \IEEEauthorrefmark{2}School of Computer Science, Wuhan University, China\\ Email: 13319269966@163.com, zouhua@whu.edu.cn, luofei@whu.edu.cn, zixian\_yang@whu.edu.cn } }

\maketitle

\begin{abstract}
Preoperative prediction of postoperative best-corrected visual acuity (BCVA) can support prognosis assessment and individualized decision-making in cataract surgery. Multimodal learning can integrate preoperative optical coherence tomography (OCT) images, scanning laser ophthalmoscopy (SLO) images, and clinical text to characterize patient status. However, SLO images are often missing in real-world clinical data due to acquisition or data-management issues, resulting in incomplete multimodal inputs. To address this problem, this paper proposes a conditional diffusion-based method for SLO proxy representation recovery and fusion prediction. Instead of reconstructing raw SLO images, the proposed method formulates SLO missingness as proxy representation recovery in the latent feature space. Specifically, OCT image features and clinical text features are first fused to construct an OCT-text conditional representation, which guides the reverse denoising process of the diffusion model to estimate sample-specific SLO proxy representations. The recovered SLO proxy representation is then used as a substitute for the unavailable SLO modality and incorporated into the downstream multimodal fusion prediction network for postoperative BCVA prediction. Experiments were conducted on 2,141 preoperative multimodal cataract samples, with an SLO missing rate of 58.80\%. The proposed method achieved an accuracy of 0.9714 under the criterion of an absolute prediction error within 0.1 LogMAR and a mean absolute error of 0.0423, outperforming representative postoperative visual acuity prediction and missing-modality learning methods. These results indicate that representation-level proxy recovery can alleviate the impact of SLO modality missingness on postoperative BCVA prediction.
\end{abstract}


\section{Introduction}

Cataract is one of the leading causes of visual impairment worldwide, and cataract surgery remains the most effective treatment for visual restoration. Postoperative best-corrected visual acuity (BCVA) is a key indicator evaluating surgical prognosis, but visual recovery after surgery still varies substantially across patients. This variability is not solely determined by lens opacity, but is also influenced by retinal structural conditions, preoperative visual function, concomitant fundus diseases, and patient-specific clinical factors \cite{bourne2021causes}. Therefore, accurate preoperative prediction of postoperative BCVA is of great importance for prognosis assessment, patient counseling, and individualized surgical decision-making \cite{kansal2018interocular,ting2019deep}.

Deep learning has been widely investigated for ophthalmic image analysis and visual acuity prediction. Existing studies have employed optical coherence tomography (OCT) images, fundus images, ensemble learning strategies, and Transformer-based architectures to estimate visual acuity or predict postoperative visual outcomes \cite{wei2021oct,lee2022estimation,kim2022deep,wang2022ctt}. However, single-modality approaches or models can only partially characterize a patient's preoperative condition. Multimodal learning provides a promising framework for integrating complementary information from preoperative OCT images, scanning laser ophthalmoscopy (SLO) images, and clinical text. Specifically, OCT images depict cross-sectional retinal morphology, SLO images provide en face fundus information, while clinical text contains preoperative visual acuity, demographic characteristics, and clinical diagnoses. Such complementary information enables a more comprehensive representation of patient status, thereby motivating multimodal postoperative BCVA prediction rather than reliance on a single information source.

Despite the promise of multimodal learning, real-world clinical datasets are often incomplete. In particular, SLO images may be unavailable due to variations in acquisition protocols, image quality control failures, or data-management issues, making it challenging to obtain complete OCT, SLO, and clinical text inputs for every sample. In the dataset used in this study, OCT images and clinical text are available for all samples, whereas SLO images are absent in a substantial proportion of cases. Therefore, effectively modeling and compensating for missing SLO information is a critical challenge for robust multimodal postoperative BCVA prediction.

Several studies have investigated multimodal representation learning and model robustness under incomplete modality settings. Ngiam et al. explored shared representation learning across multiple modalities \cite{ngiam2011multimodal}, while ModDrop improved multimodal robustness by randomly dropping modality channels during training \cite{neverova2016moddrop}. SMIL further addressed multimodal learning under severe modality missingness, and recent prompt-based or parameter-efficient approaches have explored how to adapt multimodal models to missing-modality scenarios \cite{ma2021smil,lee2023multimodal,reza2025robust}. In medical imaging, HeMIS introduced hetero-modal representation learning to handle missing MRI modalities \cite{havaei2016hemis}. For postoperative visual acuity prediction in cataract surgery, Zhou et al. proposed a masked self-attention framework to suppress unnecessary cross-modal interactions caused by missing modalities \cite{zhou2023iml}, while UML further employed modality-specific feature extraction and missing-modality mask fusion to address uncertainty in modality availability \cite{yang2025uml}. Although these methods improve robustness and preserve prediction capability under incomplete multimodal inputs, they mainly focus on reducing the negative influence of missing modalities during fusion or learning modality-invariant representations. They generally do not explicitly recover sample-specific, task-relevant representations for the unavailable SLO modality. This leaves an important gap for postoperative BCVA prediction, where the missing SLO modality may contain complementary fundus information beneficial to individualized prognosis estimation.

To address the above challenge, this study proposes a conditional diffusion-based framework for SLO proxy representation recovery and fusion-based postoperative BCVA prediction. Diffusion models have demonstrated strong capabilities in generative modeling and restoration tasks \cite{ho2020ddpm,song2020ddim,yang2023diffusion}. Instead of reconstructing raw SLO images, we formulate missing SLO handling as a representation-level recovery problem in the latent feature space. This design avoids the difficulty of generating complex retinal images and directly recovers task-relevant proxy features for the downstream prediction network. Specifically, the conditional diffusion recovery module is pretrained using samples with available SLO images, with preoperative OCT image features and clinical text features serving as conditions to guide the estimation of sample-specific SLO proxy representations. During inference, the diffusion model starts from noise in the SLO latent space and progressively denoises it under the guidance of the available OCT image features and clinical text features to recover the SLO proxy representation. The recovered proxy representation is then used as a substitute for the unavailable SLO modality and fused with available OCT and clinical text features for postoperative BCVA prediction.

In summary, the main contributions are listed as follows:
\begin{itemize}
 \item We propose a representation-level missing-modality recovery paradigm for postoperative BCVA prediction in cataract surgery. Instead of relying on zero filling, simple feature imputation, or raw SLO image reconstruction, the unavailable SLO modality is modeled as a latent proxy representation. This design shifts missing-SLO compensation from the image/input space to the task-oriented feature space, thereby avoiding the complexity of generating retinal images while directly recovering supplementary fundus-related information for multimodal BCVA prediction.
\item We develop a conditional diffusion-based SLO proxy representation recovery method. Specifically, OCT image features and clinical text features are fused to construct an OCT-text conditional representation, which guides the reverse denoising process of the diffusion model to estimate sample-specific SLO proxy representations. The recovered SLO proxy representation is then used as a substitute for the unavailable SLO modality and incorporated into the downstream multimodal fusion prediction network.
\item We validate the proposed method on 2,141 preoperative multimodal cataract samples with an SLO missing rate of 58.80\%. Experimental results show that the proposed method achieves an accuracy of 0.9714 under the criterion of an absolute prediction error within 0.1 LogMAR and a mean absolute error of 0.0423, outperforming representative postoperative visual acuity prediction and missing-modality learning methods. Ablation studies further demonstrate the effectiveness of representation-level proxy recovery for alleviating the impact of SLO modality missingness on postoperative BCVA prediction.
\end{itemize}

\section{Method}

We propose a conditional diffusion-based framework for SLO proxy representation recovery and fusion prediction under incomplete modality settings. In postoperative BCVA prediction after cataract surgery, SLO images provide complementary en face fundus information to OCT images but are often unavailable in real-world clinical data. To alleviate the impact of SLO modality missingness, the proposed framework estimates a sample-specific SLO proxy representation in the latent feature space using a conditional diffusion module, which is then combined with available OCT image features and clinical text features for postoperative BCVA prediction.

\begin{figure*}[t]
    \centering
    \includegraphics[width=0.95\textwidth]{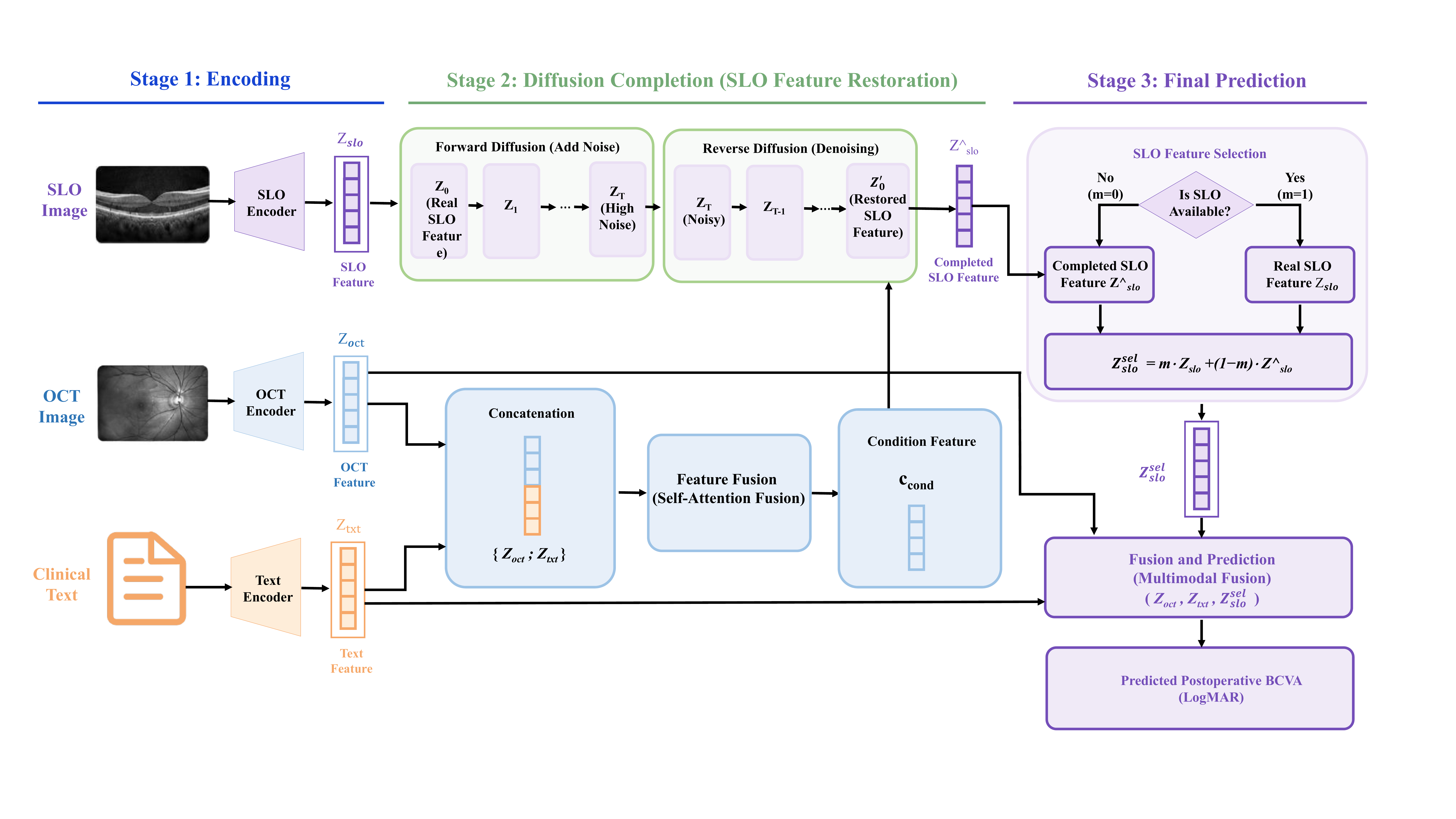}
    \caption{Overall framework of the proposed SLO proxy representation recovery and fusion prediction method. The framework consists of base prediction network training, conditional diffusion-based SLO proxy representation recovery, and final BCVA prediction with the recovered SLO representation.}
    \label{fig:framework}
\end{figure*}


\subsection{Overall Framework}

The overall workflow of the proposed method is illustrated in Fig.~\ref{fig:framework}. It contains three stages: base prediction network training, SLO proxy representation recovery pretraining, and prediction with the recovery module. In the first stage, the base network extracts OCT image features, clinical text features, and available SLO image features, and uses them for postoperative BCVA prediction through modality-specific prediction branches. This stage establishes the prediction backbone and provides modality representations for subsequent proxy recovery and baseline comparison.

In the second stage, OCT image features and clinical text features are used as conditional inputs to pretrain the SLO proxy representation recovery module. The objective is not to reconstruct raw SLO images, but to estimate a latent proxy representation that supplements the missing SLO modality. OCT images mainly describe cross-sectional retinal structures, whereas SLO images provide complementary en face fundus information. Meanwhile, clinical text contains patient-level information and clinically relevant diagnostic descriptions. Therefore, conditioning on OCT and text features enables the recovery module to infer SLO-related complementary information from the available modalities.

In the final stage, the pretrained recovery module is integrated into the main prediction network. If the SLO modality is available, the model uses the representation extracted from the real SLO image. Otherwise, it estimates the corresponding SLO proxy representation from OCT image features and clinical text features, and uses it as a substitute for the unavailable SLO modality. The real or recovered SLO representation is then combined with the available modality features for final postoperative BCVA prediction. In this way, samples with missing SLO images can still benefit from SLO-related information through representation-level proxy recovery.

We use UML~\cite{yang2025uml} as the backbone prediction network; unless otherwise specified, ``baseline'' refers to this backbone without the proposed diffusion-based SLO proxy recovery module.

\subsection{SLO Proxy Representation Recovery Module}
\label{sec:slo_recovery}

\begin{figure*}[t]
    \centering
    \includegraphics[width=0.9\textwidth]{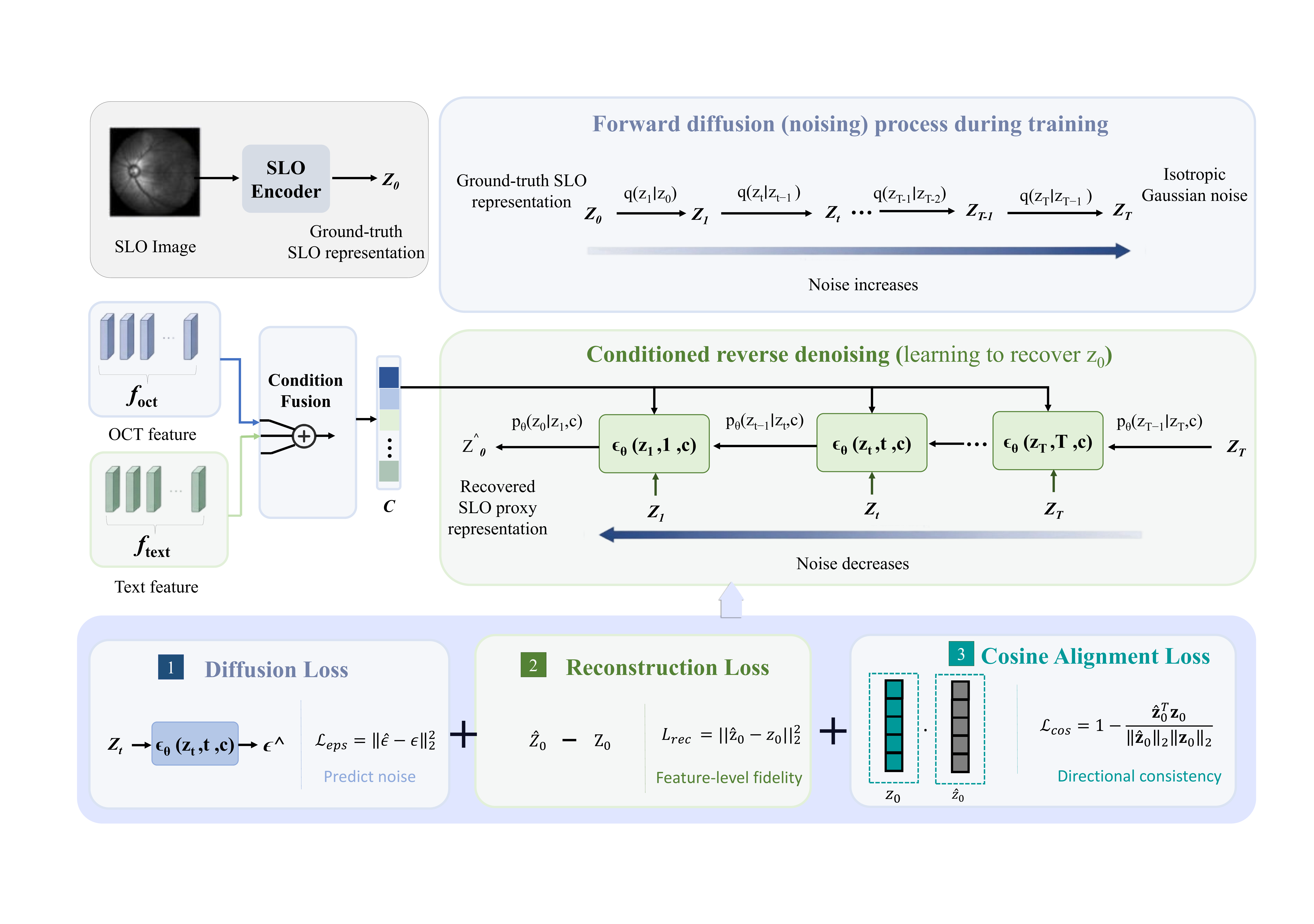}
    \caption{Conditional diffusion-based SLO proxy representation recovery module.}
    \label{fig:recovery}
\end{figure*}

The SLO proxy representation recovery module adopts a conditional diffusion model to recover SLO-related representations from available OCT image and clinical text features. As shown in Fig.~\ref{fig:recovery}, the module is pretrained only on modality-complete samples, where each sample contains an OCT image, clinical text, and an SLO image. For the $i$-th modality-complete sample, the OCT image, clinical text, and SLO image are denoted as $x_i^{oct}$, $x_i^{txt}$, and $x_i^{slo}$, respectively. Their latent representations are extracted by modality-specific encoders:
\begin{equation}
z_i^{oct}=E_{oct}(x_i^{oct}), \quad
z_i^{txt}=E_{txt}(x_i^{txt}), \quad
z_i^{slo}=E_{slo}(x_i^{slo}),
\end{equation}
where $E_{oct}(\cdot)$, $E_{txt}(\cdot)$, and $E_{slo}(\cdot)$ denote the OCT image encoder, clinical text encoder, and SLO image encoder, respectively. Accordingly, $z_i^{oct}$, $z_i^{txt}$, and $z_i^{slo}$ denote the OCT image feature, clinical text feature, and real SLO representation. In this stage, $z_i^{oct}$ and $z_i^{txt}$ are used as conditional information, while $z_i^{slo}$ serves as the recovery target.

During Stage 2 pretraining, the OCT and SLO encoders are frozen, while the precomputed clinical text representations are treated as fixed inputs. The conditional diffusion recovery components are optimized using samples with available SLO images.

Since OCT images mainly reflect cross-sectional fundus structures, while clinical text contains patient basic status, preoperative visual acuity, and related clinical information, the two modalities provide conditional constraints for SLO proxy representation recovery from structural and semantic perspectives. Therefore, OCT image features and clinical text features are first fused to construct the OCT-text conditional feature used in the reverse diffusion process:
\begin{equation}
c_i=\Phi([z_i^{oct};z_i^{txt}]),
\end{equation}
where $[z_i^{oct};z_i^{txt}]$ denotes the concatenation of OCT image features and clinical text features, $\Phi(\cdot)$ denotes the condition fusion mapping, and $c_i$ denotes the fused conditional feature. It should be noted that this condition fusion is not the multimodal fusion used in the final prediction stage, but is used to construct the condition for the SLO representation recovery process.

The training process of the diffusion model includes forward noising and reverse denoising. Let the real SLO representation be the initial clean representation of the diffusion process, namely $z_{i,0}^{slo}=z_i^{slo}$. In the forward diffusion process, Gaussian noise is gradually added to the real SLO representation to form intermediate states under different noise intensities. The SLO representation after adding noise at step $t$ can be expressed as
\begin{equation}
z_{i,t}^{slo}
=
\sqrt{\bar{\alpha}_t}z_{i,0}^{slo}
+
\sqrt{1-\bar{\alpha}_t}\epsilon,
\quad
\epsilon\sim\mathcal{N}(0,I),
\end{equation}
where $t$ denotes the diffusion step, $\bar{\alpha}_t$ denotes the cumulative coefficient determined by the noise scheduling strategy, $\epsilon$ denotes random Gaussian noise, and $z_{i,t}^{slo}$ denotes the SLO representation after noising at step $t$.

In the reverse denoising process, the conditional denoising network takes the noised SLO representation $z_{i,t}^{slo}$, the diffusion step $t$, and the OCT-text conditional feature $c_i$ as inputs, and predicts the noise added in the forward process:
\begin{equation}
\hat{\epsilon}
=
\epsilon_{\theta}(z_{i,t}^{slo},t,c_i),
\end{equation}
where $\epsilon_{\theta}(\cdot)$ denotes the conditional denoising network and $\hat{\epsilon}$ denotes the predicted noise. According to the predicted noise, the corresponding clean SLO representation is then estimated as
\begin{equation}
\hat{z}_{i,0}^{slo}
=
\frac{
z_{i,t}^{slo}
-
\sqrt{1-\bar{\alpha}_t}\hat{\epsilon}
}{
\sqrt{\bar{\alpha}_t}
},
\end{equation}
where $\hat{z}_{i,0}^{slo}$ denotes the recovered SLO proxy representation. Thus, SLO proxy representation recovery is modeled as an OCT-text-conditioned reverse denoising process in the SLO feature space.

To optimize the conditional diffusion recovery process, we first use a noise prediction loss to constrain the denoising network's ability to estimate the forward noise:
\begin{equation}
\mathcal{L}_{eps}
=
\mathbb{E}_{i,t,\epsilon}
\left[
\left\|
\epsilon
-
\epsilon_{\theta}(z_{i,t}^{slo},t,c_i)
\right\|_2^2
\right].
\end{equation}
$\mathcal{L}_{eps}$ measures the discrepancy between the real and predicted noise. Relying only on the noise prediction loss can establish the diffusion recovery process, but it is difficult to explicitly ensure that the recovered proxy representation is sufficiently close to the real SLO representation in the feature space. Therefore, we further introduce a representation reconstruction loss:
\begin{equation}
\mathcal{L}_{rec}
=
\frac{1}{N_c}
\sum_{i=1}^{N_c}
\left\|
\hat{z}_{i,0}^{slo}
-
z_i^{slo}
\right\|_2^2,
\end{equation}
where $N_c$ denotes the number of modality-complete samples used for recovery module pretraining. This loss constrains the recovered proxy representation to be close to the real SLO representation in the feature space.

In addition, subsequent multimodal fusion prediction depends not only on feature magnitude, but also on the directional relationship between features in the latent space. To enhance the overall directional consistency between the recovered representation and the real SLO representation, we introduce a cosine direction constraint:
\begin{equation}
\mathcal{L}_{cos}
=
\frac{1}{N_c}
\sum_{i=1}^{N_c}
\left(
1-
\frac{
\hat{z}_{i,0}^{slo}\cdot z_i^{slo}
}{
\left\|\hat{z}_{i,0}^{slo}\right\|_2
\left\|z_i^{slo}\right\|_2
}
\right).
\end{equation}
Here, $\hat{z}_{i,0}^{slo}\cdot z_i^{slo}$ denotes the inner product between the recovered SLO proxy representation and the real SLO representation, and $\|\cdot\|_2$ denotes the $\ell_2$ norm. This constraint does not require the recovered representation to copy the real SLO representation in every dimension, but encourages the two representations to maintain a similar overall direction in the latent space, thereby improving the stability of the proxy representation when it participates in subsequent fusion prediction.

In summary, the overall training objective of the SLO proxy representation recovery module can be written as
\begin{equation}
\mathcal{L}_{slo}
=
\mathcal{L}_{eps}
+
\lambda_{rec}\mathcal{L}_{rec}
+
\lambda_{cos}\mathcal{L}_{cos},
\end{equation}

where $\lambda_{rec}=0.5$ and $\lambda_{cos}=0.05$ denote the weighting coefficients of the representation reconstruction loss and the cosine direction constraint, respectively. This joint objective simultaneously constrains the diffusion denoising process, the numerical closeness of the recovered representation, and the directional consistency of the representation.

Different from direct modality filling or pixel-level SLO image reconstruction, the proposed module recovers a proxy representation in the latent feature space for downstream BCVA prediction. This formulation allows SLO-missing samples to obtain supplementary SLO-related representations without reconstructing raw SLO images. 

\subsection{Proxy Representation Integration and Fusion Prediction}

During the final prediction stage, the pretrained diffusion recovery module is kept frozen and used to estimate representations for missing SLO inputs, while the downstream prediction network is optimized for BCVA regression. The prediction network takes preoperative OCT image features, clinical text features, and SLO-related representations as inputs, and aggregates information from different sources through a multimodal fusion module to estimate postoperative BCVA.

During prediction, the SLO-related representation is selected according to the modality availability of each sample. For samples with available SLO images, the model directly uses the real SLO representation extracted by the SLO encoder. For samples without SLO images, the pretrained recovery module estimates a proxy representation from the OCT image feature and clinical text feature. To provide a unified SLO-related input under different modality states, we define
\begin{equation}
\tilde{z}_i^{slo}
=
m_i z_i^{slo}
+
(1-m_i)\hat{z}_i^{slo},
\end{equation}
where $m_i$ indicates the availability of the SLO modality. When $m_i=1$, $\tilde{z}_i^{slo}$ is the real SLO representation; when $m_i=0$, it is replaced by the recovered proxy representation. Therefore, $\tilde{z}_i^{slo}$ serves as the unified SLO-related representation for the prediction network.

Through the above replacement mechanism, without changing the main prediction objective or the fusion network structure, the proposed method transforms the SLO missingness problem into proxy representation completion in the feature space. Compared with directly discarding the SLO modality or using zero-value filling, this strategy provides a more informative supplementary representation when SLO is missing, allowing the final prediction network to still use potential information related to the SLO modality. Finally, the unified SLO-related representation $\tilde{z}_i^{slo}$ is fed into the multimodal fusion module together with OCT image features and clinical text features for postoperative BCVA prediction.

\subsection{Optimization Objective}

The optimization objective of the proposed method consists of the SLO proxy representation recovery objective and the BCVA main prediction objective. For the SLO proxy representation recovery module, we pretrain it on modality-complete samples and constrain it using the recovery loss $\mathcal{L}_{slo}$ defined in Section~\ref{sec:slo_recovery}. This objective guides the model to recover a proxy representation corresponding to the real SLO representation from OCT image features and clinical text features, so that it can provide supplementary information for the main task when the SLO modality is missing.

In the BCVA main prediction stage, the model feeds OCT image features, clinical text features, and the unified SLO-related representation $\tilde{z}_i^{slo}$ into the multimodal fusion prediction network, and outputs the postoperative BCVA prediction $\hat{y}_i$ for the $i$-th sample. Let the ground-truth BCVA label be $y_i$. The main prediction loss is defined as
\begin{equation}
\mathcal{L}_{bcva}
=
\frac{1}{N}
\sum_{i=1}^{N}
\left\|
\hat{y}_i-y_i
\right\|_2^2,
\end{equation}
where $N$ denotes the number of training samples, $\hat{y}_i$ denotes the predicted postoperative BCVA, and $y_i$ denotes the corresponding ground-truth label. This loss directly constrains the final prediction output, enabling the fused multimodal representation to serve the postoperative BCVA prediction task.

Therefore, $\mathcal{L}_{slo}$ and $\mathcal{L}_{bcva}$ correspond to the two core objectives of the proposed method. The former is used to learn proxy representation recovery for the missing SLO modality, while the latter is used to optimize the final postoperative visual acuity prediction. With this design, the recovery module does not act as an independent modality generation module, but is integrated into the main prediction framework so that the recovered SLO proxy representation ultimately serves BCVA prediction.

\section{Experiments}
\subsection{Experimental Settings}

We evaluate the proposed method on a multimodal dataset consisting of 2,141 preoperative cataract samples, including OCT images, SLO images, and clinical text, with postoperative best-corrected visual acuity (BCVA) as the prediction target. Each sample contains OCT images, clinical text, and BCVA labels, while SLO images are available for 882 samples and missing for 1,259 samples, corresponding to a missing rate of 58.80\%. The dataset is divided into training, validation, and test sets with a ratio of 8:1:1, resulting in 1,712, 214, and 215 samples, respectively. Detailed statistics of SLO availability are reported in Table~\ref{tab:slo_stats}.

All images are resized and normalized before training. OCT images are resized to $256 \times 256$ and then center-cropped to $224 \times 224$, while SLO images are resized to $224 \times 224$. Following UML~\cite{yang2025uml}, patient-level clinical information, including age, sex, preoperative visual acuity, and related fundus disease information, is organized into structured sentences. A pretrained MedBert model encodes these sentences into clinical text representations, which are precomputed before training. For the SLO proxy representation recovery module, only samples with available SLO images are used for pretraining, whereas the main prediction network is evaluated under the same data split for all methods to ensure fair comparison in SLO-missing scenarios. All models are trained using the Adam optimizer for up to 100 epochs with a batch size of 64, together with learning rate decay and early stopping to mitigate overfitting. The SLO proxy recovery module is pretrained independently using a DDPM with 100 diffusion steps. The target SLO representation consists of 49 tokens with 768 feature dimensions. A linear noise schedule is used, with beta increasing from $10^{-4}$ to $0.02$. The conditional denoising network employs eight attention heads, a dropout rate of 0.1, and a 768-dimensional time embedding.

We report Acc. ($\leq$0.1 LogMAR) and Mean Absolute Error (MAE) as the main evaluation metrics. A prediction is considered correct when its absolute error relative to the ground-truth BCVA does not exceed 0.1 LogMAR. This threshold corresponds approximately to one line on a standard LogMAR visual acuity chart and provides an interpretable criterion for prediction errors, rather than a validated threshold for clinical decision-making.

\begin{table}[t]
\centering
\caption{SLO Data Statistics}
\label{tab:slo_stats}
\small
\begin{tabular}{lcccc}
\toprule
Split & Samples & Available & Missing & Missing Rate \\
\midrule
Train & 1712 & 708 & 1004 & 58.64\% \\
Validation & 214 & 101 & 113 & 52.80\% \\
Test & 215 & 73 & 142 & 66.05\% \\
All & 2141 & 882 & 1259 & 58.80\% \\
\bottomrule
\end{tabular}
\end{table}

\subsection{Compared Methods}

To evaluate the effectiveness of the proposed SLO proxy representation recovery module under missing modality scenarios, we compare the proposed method with representative postoperative visual acuity prediction and missing-modality learning methods, including MAF-Net~\cite{huang2022mafnet}, CTT-Net~\cite{wang2022ctt}, MMIN~\cite{zhao2021mmin}, IML-Net~\cite{zhou2023iml}, and the UML-backbone baseline~\cite{yang2025uml}. The UML-backbone baseline and the diffusion recovery method use the same data splits, prediction backbone, and evaluation metrics, differing mainly in how missing SLO information is handled.

\textbf{UML-backbone Baseline.}
The UML-backbone baseline follows the UML framework~\cite{yang2025uml}, which uses modality-specific OCT, SLO, and text encoders with missing-modality mask fusion for postoperative BCVA prediction. In this study, it denotes the backbone prediction network without the proposed diffusion-based SLO proxy recovery module.

\textbf{Diffusion-based Proxy Recovery.}
Built upon the same UML backbone, this method replaces the original missing-SLO handling strategy with the proposed conditional diffusion-based SLO proxy representation recovery module. The recovery module estimates sample-specific SLO proxy representations from OCT image features and clinical text features, and the recovered representations are then used as substitutes for missing SLO inputs in the downstream prediction network.

\subsection{Comparison Results}

Table~\ref{tab:comparison_results} presents the comparison results on postoperative BCVA prediction. The compared methods include MAF-Net~\cite{huang2022mafnet}, CTT-Net~\cite{wang2022ctt}, MMIN~\cite{zhao2021mmin}, IML-Net~\cite{zhou2023iml}, and the UML-backbone baseline~\cite{yang2025uml}. Among all compared methods, the diffusion-based recovery method achieves the highest Acc. ($\leq$0.1 LogMAR) and the lowest MAE, indicating the effectiveness of recovering sample-specific SLO proxy representations for postoperative BCVA prediction.

\begin{table}[t]
\centering
\caption{Comparison results on postoperative BCVA prediction.}
\label{tab:comparison_results}
\begin{tabular}{lcc}
\toprule
Method & Acc. ($\leq$0.1 LogMAR) $\uparrow$ & MAE $\downarrow$ \\
\midrule
MAF-Net & 0.8542 & 0.1761 \\
CTT-Net & 0.8873 & 0.1682 \\
MMIN & 0.9169 & 0.1193 \\
IML-Net & 0.9431 & 0.1224 \\
UML & 0.9625 & 0.0457 \\
Diffusion recovery & \textbf{0.9714} & \textbf{0.0423} \\
\bottomrule
\end{tabular}
\end{table}

As shown in Table~\ref{tab:comparison_results}, the diffusion-based method improves Acc. ($\leq$0.1 LogMAR) from 0.9625 to 0.9714 and reduces MAE from 0.0457 to 0.0423 compared with the UML-backbone baseline. This indicates that explicitly recovering sample-specific SLO proxy representations can further improve the UML-based prediction backbone under SLO-missing scenarios.

\subsection{Ablation Studies and Analysis}

To analyze the impact of different conditioning strategies in the SLO proxy recovery module, we conduct an ablation study including OCT-only, text-only, concatenation, and fusion-based conditioning.

\subsubsection{Effect of Conditioning Strategies}

We first evaluate different conditioning settings for SLO proxy recovery, including OCT-only, text-only, concatenation, and fusion-based conditioning.

\begin{table}[t]
\centering
\caption{Ablation on Proxy Recovery and Conditioning Strategies}
\label{tab:cond_ablation}
\begin{tabularx}{\columnwidth}{X c c c}
\toprule
Method & Test Loss & Accuracy & MAE \\
\midrule
UML (w/o recovery) & -- & 0.9625 & 0.0457 \\
OCT only & 2.0370 & 0.9674 & 0.0430 \\
Text only & 1.9783 & 0.9442 & 0.0507 \\
OCT-text concat & 2.0094 & 0.9395 & 0.0496 \\
OCT-text fusion & 2.0068 & \textbf{0.9714} & \textbf{0.0423} \\
\bottomrule
\end{tabularx}
\end{table}

The UML-backbone baseline uses available real SLO features but does not perform diffusion-based proxy recovery for missing SLO inputs. In contrast, the conditional recovery variants estimate missing SLO representations under different conditioning strategies. This comparison evaluates the effect of proxy recovery rather than completely removing the SLO modality.

OCT-only conditioning improves performance over the UML-backbone baseline, indicating that OCT provides strong structural priors. In contrast, text-only conditioning leads to lower threshold accuracy and higher MAE, suggesting limited standalone predictive power under the clinical error criterion. Simple concatenation of modalities also fails to improve results, indicating that naive multimodal fusion is insufficient. The proposed fusion strategy achieves the highest Acc. ($\leq$0.1 LogMAR) and the lowest MAE, although it does not obtain the lowest test loss. This suggests that structured OCT-text conditioning is more effective for clinically meaningful threshold accuracy and absolute-error performance.

\subsubsection{Analysis of Recovered SLO Representations}

To assess the quality of SLO proxy recovery,
we compare recovered representations with their
corresponding real SLO features on 101 validation
samples with available SLO images. The recovered
representations achieve a mean cosine similarity
of 0.9583 and an MSE of 0.00118, indicating
substantial feature-level similarity.

\begin{figure}[t]
    \centering
    \includegraphics[width=0.95\columnwidth]
    {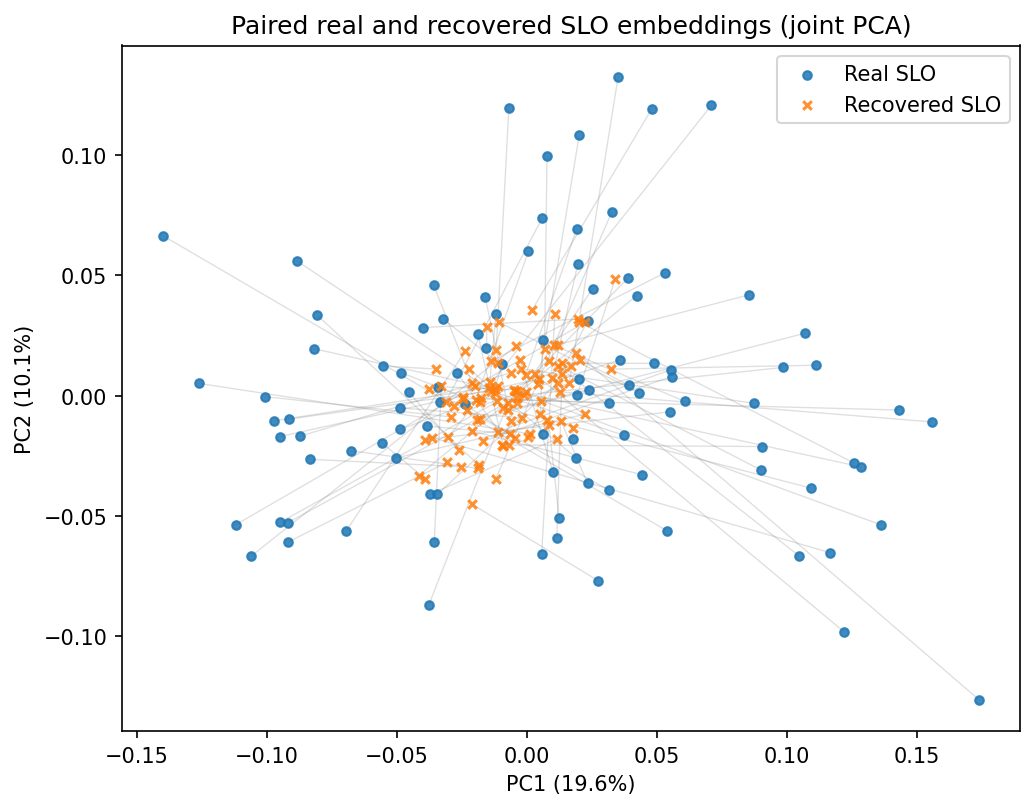}
    \caption{PCA visualization of real and
    recovered SLO representations on 101
    validation samples.}
    \label{fig:slo_pca}
\end{figure}

As shown in Fig.~\ref{fig:slo_pca}, recovered
representations exhibit a more concentrated
distribution than real SLO features, suggesting
that high representation similarity does not
necessarily imply preservation of the full
inter-sample variation.

We also measured the computational overhead
of 100-step reverse diffusion on an NVIDIA
RTX 3090 GPU. The average sampling time was
171.15 ms for a batch of two samples, equivalent
to an amortized 85.58 ms per sample. This
measurement covers diffusion sampling rather
than complete end-to-end inference.

\subsection{Summary}

Overall, OCT-text fusion-based conditional recovery achieves the best Acc. ($\leq$0.1 LogMAR) and MAE among the evaluated conditioning strategies, although it is not uniformly optimal across all metrics. The ablation studies further demonstrate that missing modality completion is not equivalent to simply adding more modalities; the way conditional information is organized has a direct impact on downstream prediction behavior. The effectiveness of recovered representations should ultimately be evaluated based on their contribution to downstream prediction tasks.

\section{Conclusion}
This paper proposed a conditional diffusion-based SLO proxy representation recovery method for postoperative BCVA prediction with missing-SLO modality. Instead of reconstructing raw SLO images or relying on simple placeholder filling, the proposed method recovers sample-specific SLO proxy representations in the latent feature space using OCT image features and clinical text features as conditional guidance. The recovered proxy representation is then incorporated into the multimodal fusion network to compensate for unavailable SLO information. Experiments on a preoperative multimodal cataract dataset with a high SLO missing rate demonstrated that the proposed method outperformed representative postoperative visual acuity prediction and missing-modality learning methods, including the UML-backbone baseline. Ablation studies further verified the effectiveness of OCT-text conditioned recovery for improving downstream prediction performance. These results suggest that representation-level proxy recovery is an effective strategy for alleviating the impact of SLO modality missingness in multimodal ophthalmic prognosis prediction.

This study is limited by its single-center dataset, the naturally varying SLO missing rates across data splits, and the absence of external validation. Further evaluation on independent clinical cohorts and under controlled missingness settings is needed to establish broader generalizability.

\bibliographystyle{IEEEtran}
\bibliography{refs}
\end{document}